%% file: 00_main.tex
\documentclass[a4paper]{styles/svproc}

\usepackage{graphicx}
\usepackage{caption}
\usepackage{subcaption}
\usepackage{multirow}
\usepackage{tabularx}
\usepackage{float}
\usepackage{tipa}
\usepackage[hidelinks]{hyperref}
\usepackage{xcolor}
\usepackage{todonotes}
\usepackage[utf8]{inputenc}
\usepackage{listings}
\usepackage{algorithm}
\usepackage{algpseudocode}
\usepackage{amsmath}
\usepackage{tikz}
\usepackage{pgfplots}
\usepackage{siunitx}
\usepackage{tabularx}
\usepackage{booktabs}
\usepackage[inline]{enumitem}
\usepackage{microtype}

\definecolor{codegray}{rgb}{0.5,0.5,0.5}
\definecolor{backcolour}{rgb}{0.98,0.98,0.98}

\graphicspath{{figures/}}

\begin{document}

\mainmatter 

\title{Towards Stress-Aware Sentence-Level Filipino G2P With Weakly-Supervised ByT5 Fine-Tuning}
\titlerunning{Stress-Aware Filipino G2P}

\author{Lorenz Bernard Marqueses \and 
        Paulo Grane Gabriel Silva \and
        Chastine Cabatay \and
        Ericson Adler Tan \and
        Ann Franchesca Laguna}

\authorrunning{L. B. Marqueses et al.}

\institute{De~La~Salle~University, Manila, Philippines, \\ \email{lorenz\_marqueses@dlsu.edu.ph}}

\maketitle

\begin{abstract}
Grapheme-to-phoneme conversion (G2P) refers to the task of converting a sequence of graphemes to a corresponding sequence of phonemes. While Filipino G2P is fairly straightforward due to its shallow orthography, the inclusion of prosodic features such as stress adds a layer of complexity that requires sentence-level context instead of single-word inputs.
However, sentence-level data for Filipino typically do not include phoneme transcriptions, posing a challenge for training G2P models. As such, we investigate how to obtain sentence-level phoneme data for Filipino using available data and compare the resulting models with multilingual word-level G2P as well as measure how accurately they predict stress marker position for Filipino.
We propose fine-tuning a ByT5-based model, pre-trained on multilingual word-level G2P data, on three sentence-level G2P datasets annotated with an LLM-assisted pipeline guided by data from Wiktionary. This approach produces models that perform well on the G2P task, achieving at best around $0.54\%$ PER and $2.50\%$ CER, a significant decrease compared to base model PER at around 19.74\%, on a manually-corrected test set. The model is able to correctly classify most of the main stress classes in Filipino, but struggles particularly with \textit{malumi} words.
We show that a ByT5-based model performs well at sentence-level Filipino G2P and offers strong potential for Filipino homograph disambiguation.
\end{abstract}


\section{Introduction}
\input{01_intro}

\section{Related work}
\input{02_related}

\section{Methodology}
\input{03_methodology}

\section{Experimental setup}
\input{04_experimental}

\section{Results and discussion}
\input{05_results}

\section{Conclusion and future work}
\input{06_conclusions.tex}

\bibliographystyle{IEEEtran}
\bibliography{refs}

\end{document}

%% file: 01_intro.tex
Automated grapheme-to-phoneme (G2P) conversion refers to the sequence transduction task where an input sequence of orthographic symbols (\textit{graphemes}) is transformed into an output sequence of phonetic symbols. For instance, the written-language sentence \textit{I will study} may be converted to the phoneme sequence \textipa{/aI "wIl "st\textturnv d.i/}. This conversion is often a crucial component when bridging the gap between written and spoken language, particularly with applications such as text-to-speech (TTS) systems. While G2P has historically been dominated by rule-based systems \cite{mortensen_epitran_2018}, advances in deep learning techniques have since highlighted the challenges posed by diverse phonologies present in different languages, as certain orthographic combinations in one language may have different corresponding phoneme patterns in others.
%
%
Filipino in particular has a shallow orthography \cite{ocampo_2004}---a characteristic it inherited from Tagalog---which means that the relationship between pronunciation and spelling is fairly direct; e.g., the word \textit{mare} would be transcribed as \textipa{/mE\textrhookschwa/} in American English, but as \textipa{/"ma\textipa{R}e/} in Filipino.

Still, Filipino has some characteristics that may complicate the G2P process, particularly with the inclusion of prosodic features. One main issue is that \textit{stress} (\textit{diin}) is lexical, which means that the position of the stressed syllable can change the meaning of a word even if they otherwise share the same spelling \cite{santiago_makabagong_2003}. For example, the word \textit{bukas} can be pronounced \textipa{/"bukas/} to mean ``tomorrow" or as \textipa{/bu"kas/} to mean ``open". Because stress is not explicitly represented in common orthography (diacritics, like the acute accent in \textit{b\'{u}kas}, are usually dropped), correct transcriptions can only be determined based on the surrounding context of the input phrase or sentence. This means that a model should learn to assign the latter pronunciation given an input like \textit{bukas ang pinto} (``the door is open"). While stress position may sometimes follow predictable patterns based on characteristics like part of speech, it remains impossible to identify deterministically as agglutination can vary phoneme placements and stress locations \cite{santiago_makabagong_2003}. For instance, \textit{sulat} (``writing") is \textipa{/"sulat/} as a noun, but the addition of the prefix \textit{nag-} to form the verb \textit{nagsulat} can shift the stress to the final syllable (\textipa{/nagsu"lat/}).
%
%
The sheer amount of possible affix permutations, each associated with their own stress patterns, makes it difficult to rely on rule-based systems for stress-aware Filipino G2P. We propose a data-centric approach to overcome these issues and produce a model that performs accurate stress-aware G2P conversions at the sentence level. All source code is made publicly available at \url{https://github.com/dlsu-cse-nlp/filipino-byt5-g2p}.
%

%% file: 02_related.tex
\subsection{Approaches to G2P}

Traditional approaches to G2P often relied on manually-crafted phonetic rule sets \cite{mortensen_epitran_2018,klosowski_rule-based_2022,pine_gi2pi_2022,matogawa_japanese_2024}. In particular, Epitran in its initial release included support for 61 languages, including Tagalog \cite{mortensen_epitran_2018}, which is the basis for Filipino. Rule-based systems are usually lightweight and achieve faster processing, but are limited by explicit definitions that may not be able to handle more irregular patterns and words. 
Statistical modeling approaches improve on this by learning patterns without relying solely on explicitly defined rules. Early approaches included joint-sequence modeling such as Sequitur G2P (using $M$-gram approximation for sequence modeling, discounted expectation maximization for training, and either best-first or $N$-best search for decoding) \cite{bisani_joint-sequence_2008} and Phonetisaurus (utilizing joint $n$-grams within a weighted finite-state transducer framework) \cite{novak_phonetisaurus_2016}, both of which were state-of-the-art at the time. With the advent of deep learning, more approaches would use \textit{sequence-to-sequence} (seq2seq) neural networks---e.g., with recurrent neural networks---taking advantage of the sequential nature of the inputs and outputs to predict phonemes based on previous time steps \cite{achanta_analysis_2016,yao_sequence--sequence_2015}. Encoder-decoder LSTM was one of the leading architectures on baseline CMUDict, NetTalk, and Pronlex datasets (with many models having the advantage of being trained on large-scale American English data) \cite{yao_sequence--sequence_2015}. 

However, modern G2P has generally moved away from such approaches in favor of transformer-based architectures. Such approaches generalize better to multiple languages; e.g., \cite{vesik_one_2020} introduced ensembled transformer models paired with a self-training regimen to achieve stronger G2P results when data is scarce. Transformer-based G2P was found to significantly outperform previous attention-less recurrent approaches on the CMUDict and NetTalk baseline datasets for English \cite{yolchuyeva_transformer_2019}. Of note is the unified paradigm introduced by the Text-to-Text Transfer Transformer (T5), which treats all language processing tasks as text generation problems. G2P with T5 relies mainly on fine-tuning a language-specific T5 model pre-trained on massive amounts of data, but if no such model exists for the intended language, a universal multilingual model (e.g., multilingual T5) \cite{xue_mt5_2021} can be used instead with slight performance trade-offs \cite{rezackova_t5g2p_2024}. G2P with T5-based models achieve high precision especially when dealing with longer words and homographs \cite{rezackova_t5g2p_2024}. This is indicative of potential applications for Filipino, which, as mentioned, is agglutinative (resulting in longer words) and features a significant amount of homographs.
%
%
Effective use of transfer learning may also be applicable to low-resource contexts, able to outperform other models trained from scratch due to making use of existing ``learned" context \cite{janyoi_transformer-based_2026}.
%

\subsection{Byte-level modeling}

The emergence of token-free models has introduced a new paradigm for phonetic tasks. ByT5, a byte-level variant of the aforementioned mT5 model, operates directly on raw UTF-8 bytes \cite{xue_byt5_2022} and has found success in G2P \cite{zhu2022byt5}. By eliminating the need for complex subword or SentencePiece tokenizers, ByT5 reallocates parameters typically dedicated to massive vocabulary matrices into dense transformer layers. This design enables the model to learn a soft vocabulary effective for character-to-phoneme mapping. In massively multilingual benchmarks, ByT5 significantly outperforms token-based models, with a ByT5-small model achieving a PER of 8.8\% versus 11.9\% for a parameter-matched mT5-small model \cite{zhu2022byt5}. Byte-level modeling further retains the capability for stronger multilingual generalization through joint learning of multiple languages; for instance, CharsiuG2P \cite{zhu2022byt5} covers G2P for around 100 languages, including Tagalog.
%
%
Even in low-resource language context, specifically for Bengali, ByT5 has been found to outperform word-based models at the G2P task \cite{islam_transcribing_2026}. Bengali shares some similar features with Filipino that pose challenges to the G2P task, specifically the usage of foreign words and variations in spelling conventions, but ByT5 is able to handle out-of-vocabulary (OOV) words and variations in word representations better than other models \cite{islam_transcribing_2026}. In the case of Brazilian Portuguese, both the multilingual CharsiuG2P model and the language-specific fine-tuned ByT5-small BIPA model with character normalization was found to perform with fairly low error rates on the BIPA dataset (6.07\% and 4.81\% minimum PER, adjusted to account for multiple possible correct transcriptions) respectively \cite{sousa-etal-2026-bipa}.
%
%
%
Furthermore, ByT5 has also been leveraged for tasks such as accent placement in Rigvedic Sanskrit, where pitch-accent marks serve as semantic and melodic units \cite{rajeev2025accent} akin to Filipino's stress-accent system.
%
%
Because ByT5 operates directly on Unicode combining marks, it provided a robust baseline for restoring prosodic information without fragile tokenization, which may be highly transferable to stress-restoration in Filipino. Despite this, the development of a stress-aware sentence-level G2P model for Filipino is mostly unexplored territory.



%% file: 03_methodology.tex
G2P can be expressed as the problem of transforming some input sequence of words $\mathbf{w} = (w_1, w_2, w_3, \dots, w_n)$ into some output sequence of phonemes $\mathbf{p} = (p_0, p_1, p_2, \dots, p_t)$ by some mapping function $f : \mathbf{w} \to \mathbf{p}$. Constructing $f$, which in this case is a neural network, thus requires a suitable training corpus comprising both word-level and sentence-level phoneme data that capture the rules of Filipino phonetics.


\subsection{Word-level phoneme data} \label{sec:wordlevel}

The basic foundation of our training data is a dictionary mapping of Filipino words to their respective pronunciations in IPA. We obtain this data through \textit{WikiPron}%
\footnote{The open-source mining tools provided by WikiPron are available at \url{https://github.com/CUNY-CL/wikipron}.}
, which sources multilingual pronunciation data from Wiktionary's Tagalog section containing the appropriate stress markers for each word. As an example, the word \textit{abot} pronounced as \textipa{/\textglotstop a"bot/} is represented as \texttt{\textglotstop a'bot}. On the other hand, if pronounced as \textipa{/"\textglotstop abot/}, it is represented as \texttt{'\textglotstop abot}. This initial run produced a total of $29,532$ unique word-pronunciation pairs, of which $20,435$ words were associated only with a single pronunciation, while $4,231$ words had multiple. We refer to the first group as \textit{non-ambiguous words} and the second as \textit{ambiguous words}. However, while these ambiguities are often due to stress location differences, they may still be semantically equivalent (e.g., due to dialectical variations or allophones). As such, we further narrow down the criteria for ambiguous words as follows:
\begin{enumerate*}[label=(\arabic*)]
  \item the word must have multiple pronunciations spanning at least two stress classes (see \autoref{tab:classification}), and
  \item these pronunciations must correspond to distinct senses or definitions in the word's Wiktionary entry.%
  \footnote{We manually scrape this data from each word's Wiktionary webpage.}
\end{enumerate*}
After collapsing duplicates into single entries, this produces $1,511$ \textit{semantically ambiguous words}. All remaining entries from the original $4,231$ are treated as non-ambiguous; in this case we take the first provided pronunciation.

The IPA annotations for the scraped data contain a total of 35 unique Unicode characters, but some of these symbols are not standard features of Filipino phonology or rarely appear in the WikiPron scrape (e.g., \textipa{/\*r/} appears only once throughout the entire set). They are thus replaced with phonemes in the Filipino phoneme inventory, largely borrowing from the native Tagalog phoneme inventory consisting of the vowels \textipa{/a, e, i, o, u/} and consonants \textipa{/p, t, k, \textglotstop, b, p, d, g, m, n, \ng, s, h, l, R, w, j/}, but also phonemes such as \textipa{/tS/} and \textipa{/dZ/}, that are particularly frequent in loanwords \cite{reid2008tagalog}. We arrive at the final inventory shown in \autoref{tab:all_phonemes}, which may be compared directly with the phoneme inventory used in \cite{ang_open_2014}; of the phonemes present in that set, we exclude the voiced palatal nasal \textipa{/\*n/}, the dental affricates \textipa{/\texttheta/} and \textipa{/D/}, and the voiceless velar fricative \textipa{/x/}, as they do not appear in the WikiPron scrape. 

\begin{table}[h]
    \centering
    \caption{Phoneme inventory of combined dataset, including sentence-level data.
        {\normalfont Compare with \cite{ang_open_2014} and \cite{reid2008tagalog}. Note the \textipa{/w/} sound may be considered both labial and dorsal.}} \label{tab:all_phonemes}
    \begin{tabularx}{\linewidth}{l|XXXX}
        \toprule
                        & \textbf{Labial} & \textbf{Dorsal} & \textbf{Coronal} & \textbf{Laryngeal} \\
        \midrule
        Plosive         & \textipa{p, b} & \textipa{k, g} & \textipa{t, d} & \textipa{\textglotstop} \\
        Nasal           & \textipa{m} & \textipa{N} & \textipa{n} \\
        Trill           \\
        Tap/Flap        & & & \textipa{R} & \\
        Fricative       & \textipa{f, v} & & \textipa{s, z, S, Z} & \textipa{h} \\
        Affricate       & & & \raggedright \textipa{\t{ts}, \t{dz}, \ \t{tS}, \t{dZ}} & \\
        Approx.         & \textipa{w} & \textipa{j} \\
        Lat. approx.    & & & \textipa{l} & \\
        \midrule
                        & \textbf{Front} & \textbf{Central} & \textbf{Back} \\
        \midrule
        Vowel           & \textipa{e, i} & \textipa{a, @} & \textipa{o, u} \\
        \bottomrule
    \end{tabularx}
\end{table}


\subsection{Sentence-level phoneme data}

\begin{figure}
    \centering
    \includegraphics[width=0.9\linewidth]{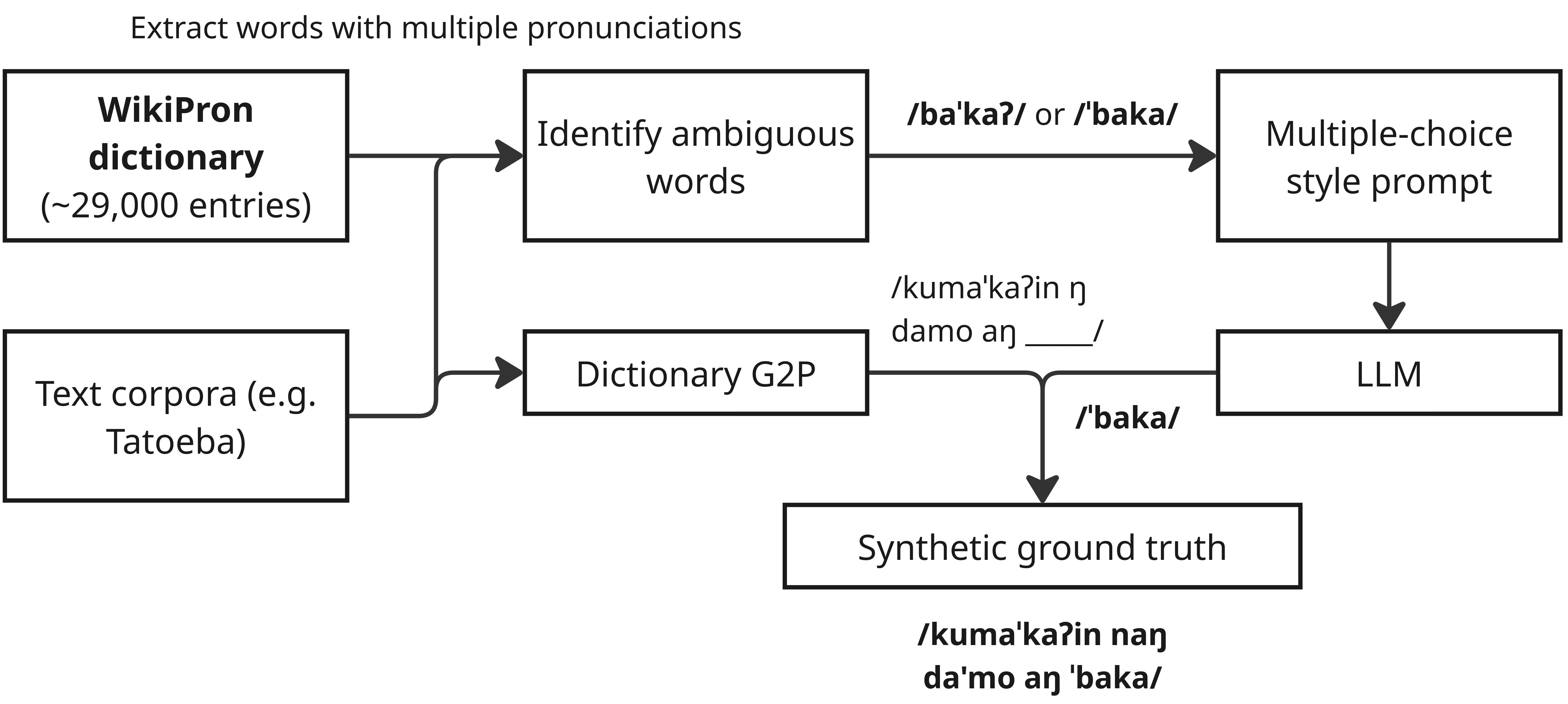}
    \caption{Synthetic training data generation pipeline.
        {\normalfont We use the Gemini series of models as the LLM, specifically Gemini 3 Flash and 2.5 Flash Lite. In some cases, we enable the LLM's ``thinking" feature, which allows it to first generate an internal chain-of-thought loop before responding (see \cite{google_gemini_thinking} for more information).}}
    \label{fig:synthetic-data}
\end{figure}

In order for the model to learn stress locations in ambiguous words, out-of-vocabulary words, and stress placement for agglutinated words, it must learn from sentence-level data that provides the necessary context surrounding a word. However, existing text corpora for Filipino specifically rarely have phoneme-level transcriptions, and even more rarely include the stress marker. Thus, we propose a straightforward LLM-assisted pipeline, shown in \autoref{fig:synthetic-data}, to synthetically generate these transcriptions, similar to the approach taken in \cite{qharabagh_fast_2025}. Each input sentence will be compared with WikiPron data to identify words with more than one possible pronunciation. We then prompt the LLM to select which of the possible pronunciations for the ambiguous word is appropriate given the context. Rule-based G2P is applied to the rest of the sentence, which at this stage should only have one possible pronunciation per word. The two outputs are then combined to produce the full phoneme-transcripted output.
An important limitation here is that WikiPron does not contain all possible inflections of each word. In these cases, the LLM will have to infer the pronunciation, so we include in the prompt the root word (obtained through the \texttt{tglstemmer} \cite{maagma_tagalog_2025} library) as a guide. Including this information in the prompt may help the LLM infer the correct pronunciation for the inflected form. Because we heavily rely on synthetically-generated noisy data, training can be said to be \textit{weakly-supervised}.

\subsubsection{Tatoeba}

The Tatoeba dataset is a collection of short sentences and their translations in various languages. The $7,300$ sentences in the original set are processed through an LLM (Gemini 3 Flash using the ``thinking" feature; see \autoref{fig:synthetic-data}), filtered, and split into a training set (roughly $80\%$), validation set (roughly $10\%$), and test set (roughly $10\%$). The filter works to remove characters that do not align with any of the phonemes in \autoref{tab:all_phonemes} (although for certain cases we map to the closest available representation of a sound, e.g. \textipa{/\textturnr/} is mapped to \textipa{/R/}), as well as to fit the generation limit set due to resource constraints. This resulted in a final count of $5,662$ training samples with a mean length of $34.9$ (SD $= 16.7$) characters per sentence. 

\subsubsection{NewsPH-NLI}

We had initially sampled $50,000$ sentences from the NewsPH-NLI dataset \cite{cruz2021exploiting}, but due to resource constraints only $31,166$ sentences were processed through the LLM (using Gemini 2.5 Flash Lite but with the same prompt as with the Tatoeba set, without ``thinking"; as the sentences were processed concurrently, this is essentially a random subset of the initial $50,000$). We apply the same filters (and mapping corrections) as with the Tatoeba set. This resulted in $22,178$ samples that are again split, producing $17,742$ training samples with a mean length of $177.6$ (SD $= 43.6$) characters per sentence (thus this is also the most complex of the available datasets, but is also the noisiest).

\subsubsection{Synthetically generated datasets}

To ensure that all ambiguous words in the WikiPron scrape are useful at the sentence level, we initially took the entire initial set of $4,231$ ambiguous words, which includes those with no meaningful semantic differences, and used an LLM to generate 5 sentences per pronunciation per word (using Gemini 3 Flash without ``thinking," with the accompanying instruction that pronunciations with no difference in meaning are collapsed into one). The result was fed as the input to the same annotation pipeline as in \autoref{fig:synthetic-data} and the same filters as the other datasets to produce a total of $27,494$ samples, of which $21,993$ are included in the training split (though only $17,484$ are generated from ambiguous words). We refer to this dataset as the \textit{naive synthetic} dataset. However, as the name implies, this involved having the LLM essentially guess the correct pronunciation for each word. For a more guided approach, we take the $1,511$ semantically ambiguous words (see \autoref{sec:wordlevel}) and generate a second synthetic dataset consisting of 20 sample sentences for each distinct pronunciation of each word (i.e., different stress classes). In this case, the LLM also receives the possible definitions of each word. It is instructed to split the 20 generated sample sentences evenly across all definitions that are associated with each pronunciation. For instance, 40 sentences total are generated for the word \textit{baka}, split between \textipa{/"baka/} and \textipa{/"baka\textglotstop/}. The earlier group comprises 4 sentences each for the senses ``declaration of war," ``campaign," ``beef," ``cattle," and ``verbal assault," but the latter group only comprises only 10 sentences each for the senses ``might" and ``maybe, probably". The generated data is then passed through the usual filters to produce a total of $59,252$ samples, of which $47,402$ were included in the training split. We refer to this dataset as the \textit{Wiktionary-guided synthetic} dataset.

\subsubsection{Manually-corrected testing set}

 We randomly sample sentences from the Tatoeba and synthetic datasets, then hand-pick and manually correct sentence transcriptions to produce a gold-standard evaluation set with a total of $1,173$ samples. We exclude NewsPH-NLI due to the relatively poor quality of its automated transcriptions that require significantly more manual corrections to be usable as an evaluation set.

\subsection{Model architecture}

\begin{figure}
    \centering
    \includegraphics[width=0.9\linewidth]{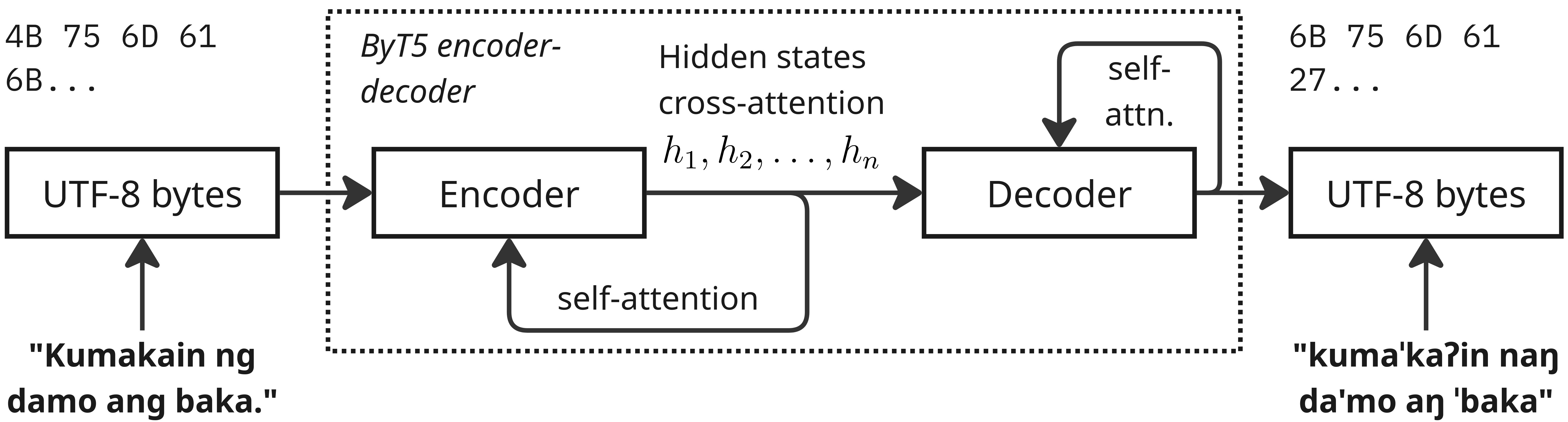}
    \caption{Model architecture overview}
    \label{fig:model-arch}
\end{figure}

As shown in \autoref{fig:model-arch}, the model uses an encoder-decoder architecture initialized with the weights from CharsiuG2P \cite{zhu2022byt5} and fine-tuned using the above combined corpus. The use of the ByT5 architecture means that input sentences and words are represented purely in UTF-8 bytes. 


%% file: 04_experimental.tex
\subsection{Evaluation metrics}

\subsubsection{Phoneme (PER) and character error rate (CER)}

The primary metrics used for model performance are the \textit{phoneme} (PER) and \textit{character error rates} (CER). The former breaks each input string into phonemes using the segmentation tool provided by PanPhon \cite{mortensen_panphon_2016}, while the latter is calculated directly using raw characters. As such, where PER only considers the mapping from letters to phonemes, CER includes whitespace and the stress marker. More precisely, for each sample $i \in \{1, 2, \dots, N\}$, both PER and CER are calculated by obtaining the Levenshtein edit distance $d$ between the ground-truth $y_i$ and predicted sequence $\hat{y_i}$, then normalizing by the total number of phonemes (or characters):
\begin{equation} \label{eq:per-cer}
    \text{PER/CER} = \frac{\sum_{i=1}^N d(\hat{y_i}, y_i)}{\sum_{i=1}^N |y_i|}
    = \frac{\sum_{i=1}^N (S_i + D_i + I_i)}{\sum_{i=1}^N |y_i|} 
\end{equation}
where $S_i$ (substitutions), $D_i$ (deletions), and $I_i$ (insertions) refer to the total number of operations required to form $y_i$ given $\hat{y_i}$, obtained through a minimum-cost alignment. 

\subsubsection{Phonetic feature error rate (PFER)}

Another measure of transcription accuracy provided in the PanPhon toolkit is the \textit{phonetic feature error rate} (PFER) \cite{mortensen_panphon_2016}. Unlike PER and CER, PFER measures similarity between articulatory feature vectors (e.g., mistaking \textipa{/b/} for \textipa{/p/} would incur a less harsh penalty due to both sounds being bilabial stops). Given some sequence of $k$ phonemes $\mathbf{p} = \{ p_1, p_2, \dots, p_k \}$, PanPhon calculates an articulatory feature vector $V(p_i)$ for each entry to produce $\text{feat}(\mathbf{p}) = \{V(p_1), V(p_2), \dots, V(p_k) \}$. As with PER and CER, we then calculate the \textit{feature edit distance} (FED) following the same distance calculation in \autoref{eq:per-cer}, except each substitution, deletion, or insertion operation is assigned a cost ($c_S$, $c_D$, or $c_I$) calculated as follows:
\begin{align}
    c_{S}(p, \hat{p}) &= \frac{1}{2F} \sum_{j=1}^F |V(p)_j - V(\hat{p})_j| \\
    c_{I}(p) = c_{D}(p) &= \frac{1}{F} \sum_{j=1}^F w(V(p)_j), \text{ for } w(x) = \begin{cases} 1 & x \neq 0 \\ 0.5 & x = 0 \end{cases}
\end{align}
where $F$ is the dimensionality (i.e., $|V(p)|$) of the feature space, which for PanPhon is fixed to $24$.

\subsubsection{Stress classification accuracy}

Filipino has four main patterns for the location of the stressed syllable and glottal stops in a word \cite{santiago_makabagong_2003}: \textit{malumay}, \textit{mabilis}, \textit{maragsa}, and \textit{malumi}. As shown in \autoref{tab:stress-patterns}, each class places stress either at the final or penultimate syllable and ends either with an open sound or a glottal stop. Given an input phoneme transcription (e.g., \textipa{/"buhaj/}), it is thus almost entirely deterministic to identify which pattern is being followed. We do this by breaking a word into syllables (e.g. $\{\text{\textipa{"bu}}, \text{\textipa{haj}}\}$) and checking for the position of stress and any glottal stops. By applying this to each word, we are able to evaluate the G2P model treating stress modeling as a multi-class classification problem; this thus gives a metric focused specifically on Filipino prosody rather than raw phonetics. Words that do not fall under these four classes are classified as \textit{nonstandard},%
\footnote{\cite{santiago_makabagong_2003} defines a secondary stressing phenomenon that may accompany the main four classes: \textit{mariin}. This occurs in the imperfect aspect (\textit{aspektong nagaganap}) or the contemplative aspect (\textit{aspektong magaganap}) of a verb. We exclude this to focus solely on the primary stress.}
while words that are completely unstressed are classified as \textit{none}.

%
%
%

\begin{table}[h]
    \small
    \centering
    \caption{Main Filipino stress patterns. Each stress class corresponds to a diacritic, but this is often excluded in everyday use.}
    \label{tab:stress-patterns}
    \begin{tabularx}{\linewidth}{l|lllllX}
        \toprule
        \textbf{Pattern} & 
            \textbf{Word} &
            \textbf{Diacritic} &
            \textbf{Stress} &
            \textbf{Ending} &
            \textbf{IPA} &
            \textbf{Meaning} \\
        \midrule
        \textit{Malumay} &
            \textit{buhay} &
            \textit{buhay} &
            Penultimate & Open & \textipa{/'buhaj/} & ``life'' \\
        \textit{Mabilis} &
            \textit{buhay} &
            \textit{buh\'{a}y} &
            Final & Open & \textipa{/bu'haj/} & ``alive'' \\
        \textit{Maragsa} &
            \textit{paso} &
            \textit{pas\^{o}} &
            Final & Glottal stop & \textipa{/pa'so\textglotstop/} & ``flower pot'' \\
        \textit{Malumi} &
            \textit{paso} &
            \textit{pas\`{o}} &
            Penultimate & Glottal stop & \textipa{/'paso\textglotstop/} & ``scald'' \\
        \midrule
        None &
            \textit{ay} & N/A &
            None & Any & \textipa{/\textglotstop aj/} & grammar particle \\
        Nonstandard &
            \textit{agila} & N/A &
            Other & Any & \textipa{/'\textglotstop agila/} & ``eagle" \\
        \bottomrule
    \end{tabularx}
\end{table}

\subsection{Training configuration}

For all models, we use the Adafactor optimizer with a learning rate of $3 \times 10^{-4}$ and an effective batch size of 64 (due to hardware constraints this required an on-device batch size of $4$ augmented by $16$ gradient accumulation steps). We set a generation limit of $256$ for validation, which is enough to cover Tatoeba, but for longer samples like those from NewsPH-NLI we simply truncate. The learning rate is kept constant but begins with a warmup scheduler with $300$ steps (lowered to 100 steps for the smaller Tatoeba dataset and increased to 600 when including the large Wiktionary-guided synthetic dataset). Furthermore, we train for $10$ epochs and take the three checkpoints for each configuration with the best validation loss, then report results using checkpoint ensembling.

%% file: 05_results.tex

\subsection{Training dataset ablation tests}

\autoref{tab:metrics-a} shows the performance of the trained models on the manually-corrected test set. The ablation shows the effects of each additional dataset on model performance; more training data generally resulted in increased model performance with lower PER, CER, and PFER. ``Pooled" refers to error rates calculated over the entire evaluation set, while ``mean" refers to the average sentence-level error rates. All the fine-tuned models greatly outperform the base CharsiuG2P model and the rule-based Epitran \cite{mortensen_epitran_2018} model, but it should be noted these two do not predict the stress marker, which means that CER is not entirely a valid metric. The dataset combinations are numbered as follows:
\begin{enumerate*}[label=(\arabic*)]
    \item Tatoeba,
    \item NewsPH,
    \item Tatoeba + NewsPH,
    \item Naive synthetic,
    \item Tatoeba + NewsPH + Naive synthetic,
    \item Tatoeba + NewsPH + Naive and Wiktionary-guided synthetic.
\end{enumerate*}
Notably, the introduction of the Wiktionary-guided synthetic set reduced CER by about $0.3\%$.

\begin{table*}
    \caption{Training dataset ablation tests showing metrics on each test set.
        {\normalfont \textit{Pooled} metrics are calculated treating the entire corpus as a single corpus. Mean and std. are reported treating each sample separately. Note that since the base model and Epitran do not generate the stress marker, CER results for (7) and (8) are not entirely valid (but are kept, marked with $\dagger$, for completion).}}

        
    \centering
    \label{tab:metrics-a}
    \begin{tabularx}{\textwidth}{l|XXX|XXX|XXX}
        \toprule
        & \multicolumn{3}{c|}{\textbf{PER (\%)}} & \multicolumn{3}{c|}{\textbf{CER (\%)}} & \multicolumn{3}{c}{\textbf{PFER (\%)}} \\ 
        & Pooled & Mean & Std. & Pooled & Mean & Std. & Pooled & Mean & Std. \\
        \midrule
        (1) & 1.19 & 1.21 & 2.16 & 4.34 & 4.32 & 3.75 & 0.84 & 0.86 & 1.55 \\
        (2) & 5.25 & 5.04 & 5.15 & 8.31 & 8.27 & 5.67 & 3.83 & 3.68 & 4.17 \\
        (3) & 1.02 & 1.01 & 2.17 & 3.90 & 3.82 & 3.52 & 0.72 & 0.72 & 1.65 \\
        (4) & 0.75 & 0.80 & 1.93 & 3.09 & 3.12 & 3.38 & 0.55 & 0.59 & 1.53 \\
        (5) & 0.53 & 0.55 & 1.43 & 2.83 & 2.79 & 3.02 & 0.39 & 0.40 & 1.10 \\
        (6) & 0.54 & 0.56 & 1.40 & 2.50 & 2.50 & 2.94 & 0.38 & 0.41 & 1.09 \\
        (7) & 19.74 & 19.84 & 7.54 & 26.64$\dagger$ & 26.82$\dagger$ & 6.27$\dagger$ & 11.10 & 11.23 & 5.11 \\
        (8) & 11.70 & 11.57 & 5.19 & 18.77$\dagger$ & 18.83$\dagger$ & 3.97$\dagger$ & 9.08 & 8.93 & 3.96 \\
        \bottomrule
    \end{tabularx}
\end{table*}


\subsection{Wilcoxon signed-rank tests}

We perform pairwise Wilcoxon signed-rank tests to assess the significance of the differences in model performance (as measured by mean sentence-level error rates) with different combinations of fine-tuning datasets. As the eight models tested results in $\binom{8}{2} = 28$ comparisons per metric over three metrics (i.e., $84$ tests), we employ the Holm-Bonferroni Method to control the family-wise error rate at $\alpha = 0.05$ with greater statistical power than with a standard Bonferroni correction. All pairwise tests were found to be significant at this threshold across all metrics with the sole exception of (5) and (6). In this case the addition of the Wiktionary-guided synthetic dataset did not result in a statistically significant improvement in PER ($W = 7,439$, $p = 1.0000$) or PFER ($W = 7,384$, $p = 1.0000$), though there was a significant difference in CER ($W = 51,352$, $p = 0.0015$). This is consistent with what can be observed in Table 3, where (5) and (6) show nearly identical error rates. Since CER is the only metric that measures stress marker placement, this finding may be interpreted to mean that the Wiktionary-guided synthetic dataset may have helped the model distinguish between stress-minimal variants without significantly affecting general performance with phonetic transcription (PER and PFER). The remaining tests suggest that introducing additional data is generally beneficial for the model. We leave more rigorous testing of these hypotheses to future work.

\begin{table}
    \caption{Stress classification accuracy metrics, with all non-ambiguous words excluded. 
        {\normalfont Results are normalized to the support of each class.}}
    \label{tab:classification}
    \begin{tabularx}{\linewidth}{X|rrrr|rrrr}
        \toprule
        \textbf{Class} & \textbf{Prec.} & \textbf{Rec.} & \textbf{F1} & \textbf{Supp.} & \textbf{TP} & \textbf{FP} & \textbf{FN} & \textbf{TN} \\
        \midrule
        \textit{Malumay} & 0.87 & 0.89 & 0.88 & 1846.00 & 1638.00 & 255.00 & 208.00 & 2513.00 \\
        \textit{Malumi} & 0.71 & 0.83 & 0.76 & 144.00 & 120.00 & 50.00 & 24.00 & 4420.00 \\
        \textit{Mabilis} & 0.90 & 0.86 & 0.88 & 1864.00 & 1606.00 & 185.00 & 258.00 & 2565.00 \\
        \textit{Maragsa} & 0.78 & 0.83 & 0.80 & 259.00 & 215.00 & 61.00 & 44.00 & 4294.00 \\
        None & 0.96 & 0.95 & 0.95 & 448.00 & 424.00 & 18.00 & 24.00 & 4148.00 \\
        Nonstandard & 0.79 & 0.62 & 0.69 & 53.00 & 33.00 & 9.00 & 20.00 & 4552.00 \\
        \bottomrule
    \end{tabularx}
\end{table}


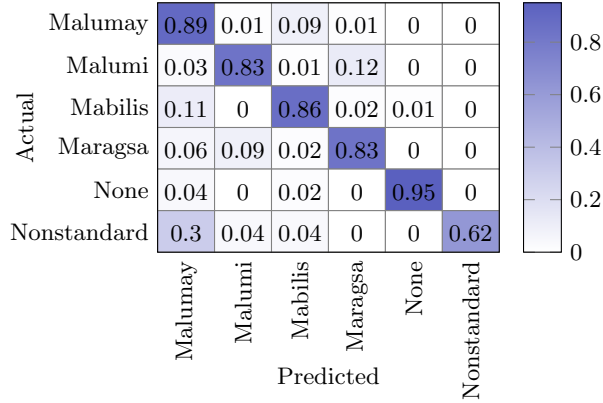
\begin{figure}
    \centering
    \begin{tikzpicture}
        \begin{axis}[
                width=0.5\columnwidth,
                height=0.4\columnwidth, 
                /pgf/number format/fixed,
                /pgf/number format/precision=2,
                colormap={bluewhite}{color=(white) rgb255=(90,96,191)},
                xlabel=Predicted,
                xlabel style={yshift=-25pt},
                ylabel=Actual,
                ylabel style={yshift=15pt},
                xticklabels={Malumay, Malumi, Mabilis, Maragsa, None, Nonstandard},
                xtick={0,...,5},
                xtick style={draw=none},
                yticklabels={Malumay, Malumi, Mabilis, Maragsa, None, Nonstandard},
                ytick={0,...,5},
                ytick style={draw=none},
                enlargelimits=false,
                colorbar,
                xticklabel style={rotate=90},
                nodes near coords={\pgfmathprintnumber\pgfplotspointmeta},
                nodes near coords style={yshift=-7pt},
            ]
            \addplot[
                matrix plot,
                mesh/cols=6,
                point meta=explicit,draw=gray
            ] table [meta=C] {
                x y C
                0 0 0.89
                1 0 0.01
                2 0 0.09
                3 0 0.01
                4 0 0.00
                5 0 0.00
                
                0 1 0.03
                1 1 0.83
                2 1 0.01
                3 1 0.12
                4 1 0.00
                5 1 0.00
                
                0 2 0.11
                1 2 0.00
                2 2 0.86
                3 2 0.02
                4 2 0.01
                5 2 0.00
                
                0 3 0.06
                1 3 0.09
                2 3 0.02
                3 3 0.83
                4 3 0.00
                5 3 0.00
                
                0 4 0.04
                1 4 0.00
                2 4 0.02
                3 4 0.00
                4 4 0.95
                5 4 0.00
                
                0 5 0.30
                1 5 0.04
                2 5 0.04
                3 5 0.00
                4 5 0.00
                5 5 0.62
            };
        \end{axis}
    \end{tikzpicture}
    
    \caption{Stress classification confusion matrix, with all non-ambiguous words excluded.
        {\normalfont Many nonstandard words are being incorrectly classified as \textit{malumay}, but the strong diagonal shows words are generally classified well.}}
    \label{fig:confusion-matrix-all}
\end{figure}

\subsection{Stress classification}

\autoref{fig:confusion-matrix-all} shows that, for each of the four main stress classes, our best model correctly classifies from 83\% to 89\% of words. It should be noted that this includes only words that we consider semantically ambiguous (meaning we measure classification performance only for cases where the model has to correctly choose one of multiple possible pronunciations from based on surrounding sentence context). Notably, in terms of standard classification metrics, the model performed the worst with \textit{malumi} words out the four main stress classifications. The model exhibits highest classification accuracy in identifying words with no stress in pronunciation. However, the model struggles with \textit{nonstandard} words as it misclassifies around 30\% of nonstandard words as \textit{malumay} instead. Classification performance is summarized in \autoref{tab:classification}. 
However, it is also noteworthy that most the errors in the test set are contributed by misplaced stress marks for homographs, such as in the following sentence, given the input ``Tulungan mo siya.", we get \textipa{/\textbf{tulu"Nan} mo si"a/} and not the expected \textipa{/\textbf{tu"luNan} mo si"a/}.


\subsection{Word-level embeddings}

\begin{figure}[H]
    \centering

    \includegraphics[width=0.8\linewidth]{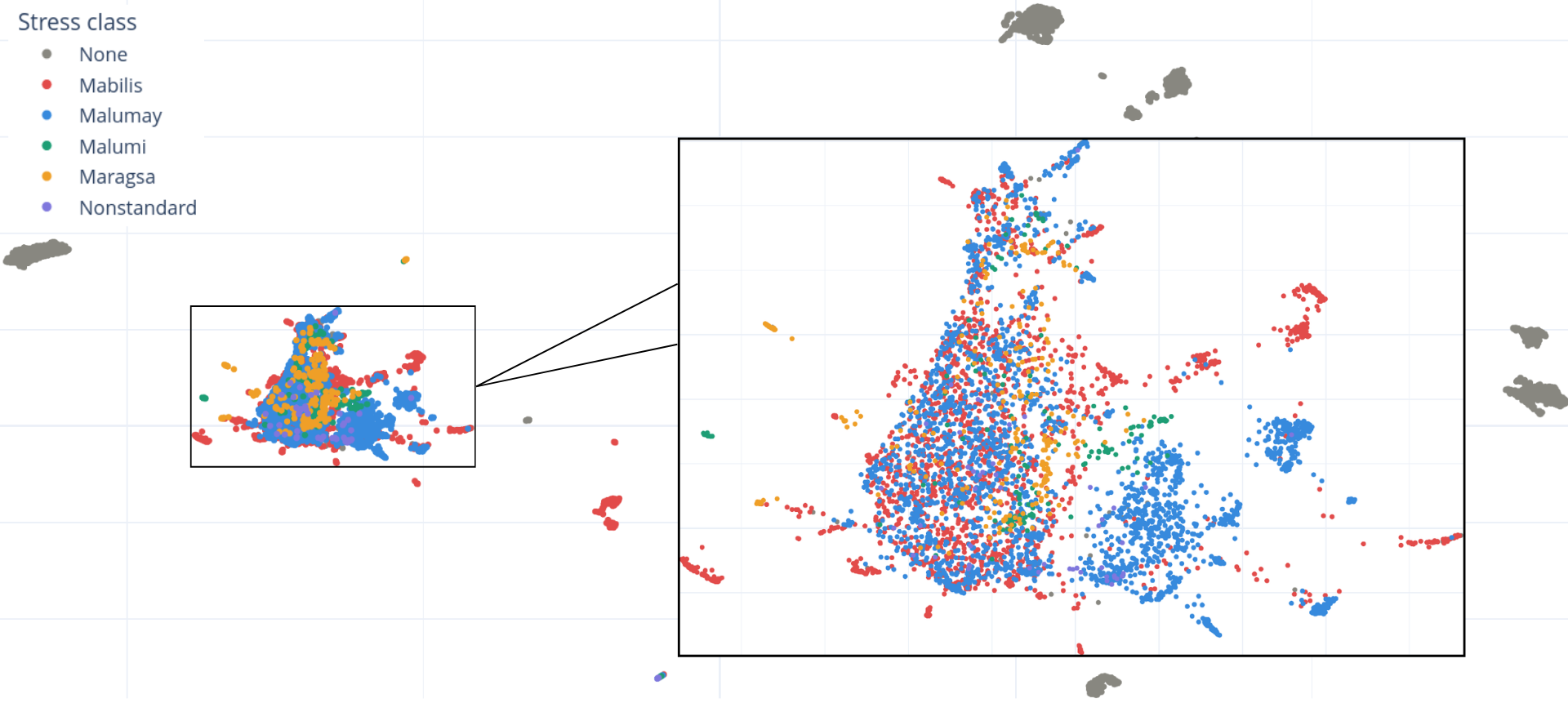}

    \caption{UMAP word-level embeddings, colored with stress classification.
        {\normalfont Byte embeddings corresponding to one word are mean-pooled to obtain a representation for a word. Observe high degree of separation for unstressed words.}}
    \label{fig:umap}
\end{figure}

We show UMAP-reduced embedding visualizations at the word level extracted from the decoder in \autoref{fig:umap}. As the ByT5 model processes bytes, we simply mean-pool the embeddings for all bytes corresponding to a word to obtain a representation for that word, ignoring cases where we are unable to find a proper alignment with the target and predicted phonemes. The visualization shows only weak clustering and very heavy overlap in stress classification in word-level embeddings, although there is a high degree of separation for unstressed words as well as small streaks for the \textit{mabilis} class. There is also a fairly distinct blob of \textit{malumay} words interspersed with \textit{mabilis} words. This may mean either the model is struggling to assign stress, too much information is compressed away with the mean-pool and the UMAP reduction, or the model simply does not explicitly represent stress location as a separable feature internally, despite learning to predict the stress marker token.

%% file: 06_conclusions.tex

%
We show that ByT5 is effective for sentence-level Filipino G2P and has potential for Filipino homograph disambiguation. We also introduce phoneme-labeled datasets based on Tatoeba and NewsPH-NLI, as well as a dataset synthetically generated through LLM-assisted pipelines. Future work must focus efforts on utilizing larger datasets for training to introduce a larger scope of diversity as well as data cleaning to ensure that training and evaluation samples use the correct stress placement for all words. We also suggest that future work look into multi-task learning to aid in predicting stress location or fusion with other models (such as the Tagalog-pretrained RoBERTa \cite{cruz2021exploiting}) to provide contextual information.
%
%
In particular, a Hybrid TagRoBERTa + Conformer \cite{gulati_conformer_2020} pipeline with an auxiliary homograph loss head may help in addressing homograph ambiguity, agglutinative verb shifts, and stress representation.